\documentclass[final]{cvpr}

\usepackage{times}
\usepackage{epsfig}
\usepackage{graphicx}
\usepackage{amsmath}
\usepackage{amssymb}
\usepackage{booktabs}
\usepackage{arydshln}
\usepackage{mathtools}
\usepackage{multirow}
\usepackage{enumitem}
\usepackage{color}
\def\red#1{\textcolor[rgb]{1,0,0}{#1}}
\def\blue#1{\textcolor[rgb]{0,0,1}{#1}}

\newcommand{\keypoint}[1]{\vspace{0.1cm}\noindent\textbf{#1}\quad}
\newcommand{\cut}[1]{}
\usepackage{algorithm}
\usepackage{algpseudocode}

\usepackage{pifont}
\newcommand{\cmark}{\ding{51}}%
\newcommand{\xmark}{\ding{55}}%

\usepackage[nopar]{lipsum}
\DeclareMathAlphabet\mathbfcal{OMS}{cmsy}{b}{n}

\makeatother
\usepackage[pagebackref=true,breaklinks=true,letterpaper=true,colorlinks,bookmarks=false]{hyperref}

\def\cvprPaperID{2304} 
\def\confYear{CVPR 2021}

\begin{document}

\title{More Photos are All You Need: Semi-Supervised Learning for \\ Fine-Grained Sketch Based Image Retrieval}
\author{Ayan Kumar Bhunia\textsuperscript{1} \hspace{.2cm} Pinaki Nath Chowdhury\textsuperscript{1,2} \hspace{.2cm} Aneeshan Sain\textsuperscript{1,2}  \hspace{.2cm}    Yongxin Yang\textsuperscript{1,2} 
\hspace{.05cm} \\ Tao Xiang\textsuperscript{1,2} \hspace{.2cm}  Yi-Zhe Song\textsuperscript{1,2} 
\\ \textsuperscript{1} SketchX, CVSSP, University of Surrey, United Kingdom \hspace{.08cm} \\
\textsuperscript{2} iFlyTek-Surrey Joint Research Centre
on Artificial Intelligence.\\
{\tt\small \{a.bhunia, p.chowdhury, a.sain, yongxin.yang, t.xiang, y.song\}@surrey.ac.uk}
}


\maketitle

\vspace{-0.2cm} 
\begin{abstract}
\vspace{-0.25cm}
A fundamental challenge faced by existing Fine-Grained Sketch-Based Image Retrieval (FG-SBIR) models is the data scarcity -- model performances are largely bottlenecked by the lack of sketch-photo pairs. Whilst the number of photos can be easily scaled, each corresponding sketch still needs to be individually produced. In this paper, we aim to mitigate such an upper-bound on sketch data, and study whether unlabelled photos alone (of which they are many) can be cultivated for performances gain. In particular, we introduce a novel semi-supervised framework for cross-modal retrieval that can additionally leverage large-scale unlabelled photos to account for data scarcity. At the centre of our semi-supervision design is a sequential photo-to-sketch generation model that aims to generate paired sketches for unlabelled photos. Importantly, we further introduce a discriminator guided mechanism to guide against unfaithful generation, together with a distillation loss based regularizer to provide tolerance against noisy training samples. Last but not least, we treat generation and retrieval as two conjugate problems, where a joint learning procedure is devised for each module to mutually benefit from each other. Extensive experiments show that our semi-supervised model yields significant performance boost over the state-of-the-art supervised alternatives, as well as existing methods that can exploit unlabelled photos for FG-SBIR.

\end{abstract}


\vspace{-0.5cm}
\section{Introduction}
\vspace{-0.1cm}
With the ever rising popularity of touch screen devices, sketch-based image retrieval (SBIR) has witnessed significant interest within the vision community \cite{bhunia2020sketch, stylemeup, sampaio2020sketchformer, zhang2018generative, dutta2019semantically, yelamarthi2018zero}. Despite starting as a category-level retrieval problem \cite{collomosse2019livesketch, dey2019doodle, sketch2vec, SBIR_imbalance}, the fine-grained nature of sketches stirred current research focus more towards fine-grained SBIR (FG-SBIR) \cite{bhunia2020sketch, pang2020solving} -- which aims to retrieve a \emph{particular} photo based on a query sketch at an intra-category basis.

\begin{figure}[]
\begin{center}
  \includegraphics[height=4.5cm, width=0.9\linewidth]{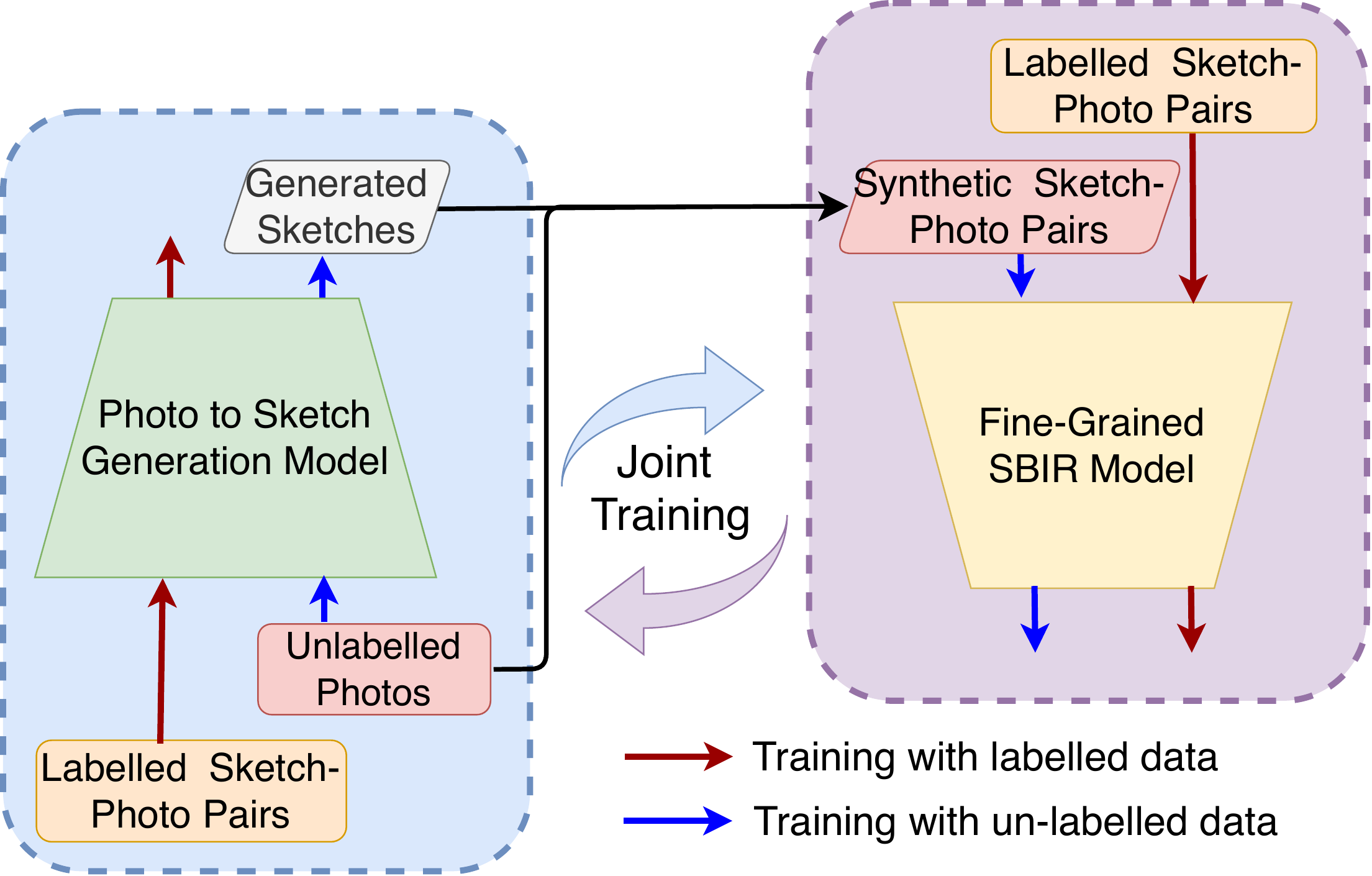}
\end{center}
\vspace{-0.4cm}
  \caption{Our proposed method additionally leverages large scale photos without any manually labelled paired sketches to improve FG-SBIR performance. Moreover, we show that the two conjugate process, \emph{photo-to-sketch} generation and \emph{fine-grained SBIR}, could improve each other by joint training.}
  \vspace{-0.7cm}
\label{fig:Fig1_a.pdf}
\end{figure}

Recent FG-SBIR works \cite{yu2016sketch, pang2019generalising, bhunia2020sketch, pang2020solving} predominately rely on \textit{fully-supervised} triplet loss-based deep networks to yield retrieval performances of practical value. The underlying assumption is largely inline with the progression of supervised photo-only models -- that one can always (relatively easily) obtain additional labelled training data to sustain desired performance gains. This assumption however does not hold for FG-SBIR -- sketch-photo pairs can not be easily scaled as per their photo-only counterparts. That is, instead of crawling and then labelling photos, the corresponding sketch for any given photo will need to be separately drawn by hand. As a result, current FG-SBIR datasets still remain in their thousands (6.7K for QMUL-ShoeV2 \cite{yu2016sketch, song2017deep}, and 2K for QMUL-ChairV2 \cite{song2017deep}), while photo-datasets \cite{russakovsky2015imagenet} are available in millions. This data scarcity problem has consequently resulted in very recent attempts that aim at designing generalisable and zero-shot models \cite{pang2019generalising}, yet performances of these models remain far away from fully-supervised alternatives. 

In this paper, we face the music and make the bold assumption that there will hardly be sufficiently large sketch-photo pairs to train a good model. Instead, we test the hypothesis that -- freely-available \textit{unlabelled photo data} would help to mitigate the performance gap imposed by the lack of specifically collected \textit{photo-sketch paired data}. Our utmost contribution is therefore a semi-supervised FG-SBIR framework where unlabelled photo data (i.e., photos without matching sketches) are used alongside photo-sketch pairs for model training. We differ significantly to conventional semi-supervised classification methods \cite{sohn2020fixmatch, pham2020meta} -- other than learning pseudo photo labels via a learnable classifier, our ``label'' for a photo is in the form of a visual sketch which needs to be \textit{generated} rather than \textit{classified}. Thus, at the core of our design is a sequential photo-to-sketch \textit{generation model} that outputs pseudo sketches for unlabelled photos. The hope is therefore that such pseudo sketch-photo pairs could augment the training of a \textit{retrieval model}.

 

Naively cascading a generator with a retrieval model however would not work. This is mainly because off-the-shelf photo-to-sketch generation models \cite{song2018learning, chen2017sketch} could sometimes generate unfaithful sketches that may not resemble their corresponding photos, especially when it comes to fine-grained visual features. The downstream retrieval model would then have no way of knowing which pseudo sketch and photo pairs are worth training with, ultimately resulting in performance degradation. This leads to an important design consideration of ours -- we advocate that there is positive complementarity between generation and retrieval that can be explored via \emph{joint learning} (Figure \ref{fig:Fig1_a.pdf}). The intuition is simple -- pseudo sketches automatically generated from unlabelled photos can help to semi-supervise a better retrieval model, \textit{and vice versa} that better retrieval model can feed back to the generator in producing more faithful sketch-photo pairs.

\begin{figure*}[t]
\begin{center}
  \includegraphics[width=\linewidth]{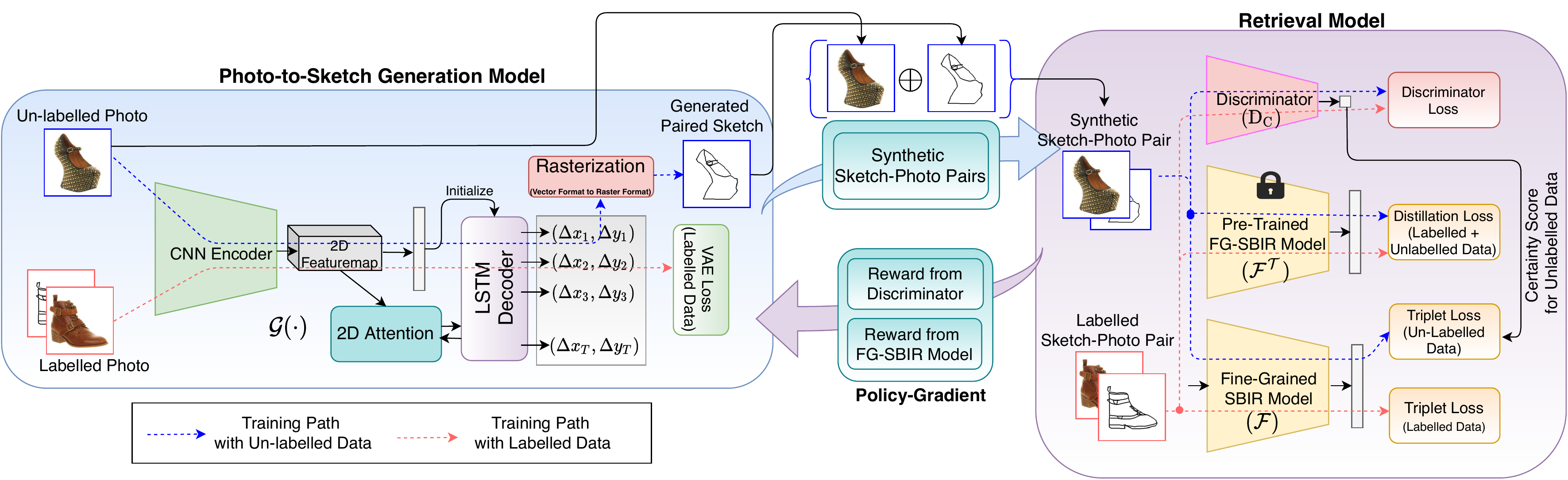}
\end{center}
\vspace{-.20in}
  \caption{Our framework: a FG-SBIR model ($\mathcal{F}$) leverages large scale unlabelled photos using a \emph{sequential}   photo-to-sketch generation model ($\mathcal{G}$) along with labelled pairs. Discriminator ($\mathrm{D_C}$) guided instance-wise weighting and distillation loss are used to guard against the noisy generated data. Simultaneously, $\mathcal{G}$ learns by taking reward from $\mathcal{F}$ and $\mathrm{D_C}$ via policy gradient (over both labelled and unlabelled) together with supervised VAE loss over labelled data. Note rasterization (vector to raster format) is a non-differentiable operation.}
\vspace{-.55cm}
\label{fig:Fig2}
\end{figure*}

The key therefore lies with \textit{how} such positive exchange cycle can be facilitated between the generator and retrieval model. 
To this end, novelty lies in the components introduced in both generator and retrieval model designs, and in how\cut{ two models} they are jointly trained.
\emph{First}, we formulate a novel sequential photo-to-sketch generator with spatial resolution preservation and a cross-modal 2D-attention mechanism. \emph{Second}, a discriminator is formulated in the retrieval model, to quantify the reliability of generated pseudo photo-sketch pairs. Reliability scores are then used\cut{to perform} for instance-wise weighting of triplet-loss values upon updating the retrieval model. A consistency loss (via distillation) is further introduced to simultaneously suppress the noisy training signal coming from pseudo photo-sketch pairs. \emph{Third}, to enable exchange from retrieval to generation, we rely on the following intuition -- good synthetic pairs would trigger a low value on the resulting triplet loss and a higher output of the discriminator. Feeding these training signals back to the generator would however involve passing through a non-differentiable rasterization operation (Figure \ref{fig:Fig2}). We thus employ a policy-gradient \cite{sutton2000policy} based reinforcement learning scheme that feeds back these signals as \textit{rewards}.

In summary, our contributions are: (a) For the first time, we propose to solve the data scarcity problem in FG-SBIR by adopting   \textit{semi-supervised} approach that additionally leverages large scale unlabelled photos to improve retrieval accuracy. (b) To this end, we couple sequential sketch generation process with fine-grained SBIR model in a joint learning framework based on reinforcement learning. (c) We further propose a novel photo-to-sketch generator and introduce a discriminator guided \emph{instance weighting} along with \emph{consistency loss}  to retrieval model training with  noisy synthetic photo-sketch pairs. (d) Extensive experiments validate the efficacy of our approach for overcoming data scarcity in FG-SBIR (Figure \ref{fig:ablation3}) -- we can already reach performances at par with prior arts with just a fraction ($\approx$60$\%$) of the training pairs, and obtain state-of-the-art performances on both QMUL-Shoe and QMUL-Chair with the same training data (by $\approx$6$\%$ and $\approx$7$\%$  respectively).

\vspace{-0.20cm}
\section{Related Works}
\vspace{-0.1cm}
\noindent \textbf{Fine-Grained SBIR:} 
Yu \etal \cite{yu2016sketch} introduced the first deep FG-SBIR model which employed a deep triplet network to learn a common embedding space for photo and sketch. Subsequent works have aimed at improving this via attention mechanisms with higher order retrieval loss \cite{song2017deep}, joint discriminative-generative learning with cross-modal image generation \cite{pang2017cross}, text tags \cite{song2017fine}, and cross-modal hierarchical co-attention \cite{BMVC_hierarchy}. Cross-category generalisation \cite{pang2019generalising}  and on-the-fly retrieval setup \cite{bhunia2020sketch} are more recent additions to existing FG-SBIR literature. \cut{As mentioned earlier, all} These fully supervised methods suffer from \textit{the} data scarcity,\cut{ problem,} which\cut{this work} we aim to address. 

\noindent \textbf{Handling \emph{Data-Scarcity} for FG-SBIR:} 
Earlier works have tried resolving the lack of instance-level photo-sketch paired data, by using edge-maps for training \cite{radenovic2018deep} or synthetic sketch stroke deformation \cite{yu2016sketchAnet, yu2016sketch} for data augmentation. Umar \etal \cite{riaz2018learning, muhammad2019goal} leveraged reinforcement-learning (RL) in an attempt to augment sketches from edge-maps under the assumption that real sketch-strokes are a subset of edge-maps, which however is negated by the highly abstracted nature of real sketches. Very recently, mixed-modality jigsaw solving \cite{pang2020solving} has been used as a pre-training task for FG-SBIR to exploit additional photo images and their edge maps for cross-modal matching. Its efficacy remains limited however as edge-maps are not sketches.

\noindent \textbf{Photo-to-Sketch Generation:} A plausible\cut{Another possible} solution to data scarcity is synthesising sketches for unlabelled photos to form pseudo photo-sketch pairs. Existing photo-to-sketch generation methods can be classified into two types: the first employs image-to-image translation \cut{framework}\cite{li2019photo}, which however merely works as a contour detection paradigm, thus failing to model the hierarchically abstracted \cut{stroke-by-stroke} nature of \cut{real} human-drawn sketch. The second follows the seminal work of Sketch-RNN \cite{ha2017neural}, and generates sequential sketch-coordinates given a photo, thus mimicking subjective human sketching style. The basic design \cite{chen2017sketch}, involving a CNN encoder and RNN decoder, has been further augmented with both self-domain and two way cross-modal reconstruction losses \cite{song2018learning}. Following this path, we improve sequential sketch-generative process with a 2D attention mechanism  to better exploit the spatial-layout of objects in photos.  


\noindent \textbf{Semi-supervised Learning:} Our learning framework is semi-supervised in the sense that the majority of training data are unlabelled photos without their paired sketches. This is thus very different from most existing semi-supervised learning methods which are designed for classification rather than cross-modal retrieval. This means that these methods, based on either  pseudo-labelling  \cite{lee2013pseudo,muller2019does,sohn2020fixmatch,pham2020meta} or consistency regularisation \cite{berthelot2019mixmatch, berthelot2020remixmatch}, offer little insight into how our problem can be solved.  
In contrast, prior works on semi-supervised cross-modal learning such as image captioning \cite{chen2016semi, kim2019image} are more relevant. However, we uniquely address a cross-modal instance-level retrieval problem, and train the generator jointly with the retrieval model, rather than merely providing model pre-training. 


\vspace{-0.2cm}
\section{Methodology}
\vspace{-0.1cm}
\noindent \textbf{Overview:} For semi-supervised fine-grained SBIR, we consider having access to a limited amount of labelled photo-sketch pairs $\mathrm{\mathcal{D}_{L} = \{ ({p_{L}^{i}}; {s_{p,L}^{i}})\}_{i}^{N_{L}}}$ and a much bigger set of unlabelled photos  $\mathrm{\mathcal{D}_{U} = \{ {p_{U}^{i}} \}_{i}^{N_{U}}}$, where $N_{U} \gg N_L$. The key objective is to improve retrieval performance using both $\mathcal{D}_{L}$ and unlabelled photos $\mathcal{D}_{U}$ (having no corresponding sketches). More specifically, our framework consists of two models learned jointly: a FG-SBIR model, and a photo-to-sketch generation model. The retrieval model tries to learn an embedding function $ \mathcal{F(\cdot)}: \mathbb{R}^{H \times W \times 3} \rightarrow \mathbb{R}^{d}$ mapping any rasterized sketch or photo having height $H$ and width $W$ to a $d$-dimensional feature.

Instead of image-to-image translation \cite{isola2017image, li2019photo}, sketch generation process needs to be designed by sequential sketch-coordinate decoding \cite{ha2017neural} in order to model the hierarchical abstract nature of sketch. In particular, the FG-SBIR model requires rasterized sketch-images to obtain the sketch embedding, as  performance can collapse on using sketch-coordinate instead \cite{bhunia2020sketch, BMVC_hierarchy}. Thus, the generator learns a function $\mathcal{G}(\cdot): \mathbb{R}^{H \times W \times 3} \rightarrow \mathbb{R}^{T\times2}$, mapping a photo to equivalent sequential sketch-coordinate points $\mathrm{s_c = \{(x_1, y_1),(x_2, y_2), \cdots, (x_T, y_T) \}}$, where $T$ is the number of points.
Note that in order to feed the generated sketches to the feature embedding network $ \mathcal{F}$ of the retrieval model,  a  \emph{rasterization} (sketch-image redrawing from coordinates) operation is required which is denoted as  $s_p = \phi(s_c) : \mathbb{R}^{T\times2} \rightarrow \mathbb{R}^{H \times W \times 3}$. Finally, we can create synthesised photo-sketch pairs  $\mathrm{\mathcal{D}_{U}' = \{ ({p_{U}^{i}}; {s_{p,U}^{i}})\}_{i}^{N_{U}}}$ for unlabelled photos to train the retrieval model. Once trained, only $\mathcal{F(\cdot)}$ is used for retrieval during inference,  while $\mathcal{G(\cdot)}$ augments pseudo/synthetic photo-sketch pairs for training.

\vspace{-0.1cm}
\subsection{Photo to Sketch Generation Model} \label{basemodel}

The existing sequential photo-to-sketch generation models \cite{song2018learning, chen2017sketch} comprise a convolutional image encoder, followed by an LSTM decoder. This however has two major limitations: Firstly, it reduces the photo representation to a latent vector, leading to significant spatial information loss. Secondly, one fixed global representation is given as input at every time step of the LSTM decoding. To overcome these limitations two novel designs are introduced: (a) keeping the spatial feature map while ignoring global average pooling; (b) looking back at the specific part of photo which it \emph{draws}. Overall, it consists of three major components, a CNN encoding the photo, a 2-D attention module, and an LSTM decoder generating the coordinates sequentially.  

Given a photo $p$, let the extracted convolutional feature map be $\mathcal{B} \in \mathbb{R}^{h'\times  w' \times c}$ where $h'$, $w'$ and $c$ denotes the height, width and number of channels, respectively. Next, we perform a global average pooling on $\mathcal{B}$ to obtain a vector of size $\mathbb{R}^c$, and project it as two vectors $\mu$ and $\sigma$, each having size $\mathbb{R}^{N_z}$. The global embedding of photo is obtained through a reparameterization trick as $z = \mu + \sigma \odot \mathcal{N}(0,1)$. The initial hidden state $h_{0}$ (and optional cell state $c_{0}$) of decoder RNN is initialised as $[h_0; c_0] = \tanh(W_{z}z + b_z)$. 

Instead of predicting the absolute coordinates $\{(x_i, y_i)\}_{i}^{T}$, we model every point as 5 element vectors $(\Delta x, \Delta y, p_1, p_2, p_3)$  where $\Delta x$ and $\Delta y$ represents the off-set distances \cite{ha2017neural} in the $x$ and $y$ directions from the previous point. The last three elements represent a binary one-hot vector of three pen-state situations: pen touching the paper, pen being lifted and end of drawing. Each offset-position ($\Delta x, \Delta y$) is modelled  using a  Gaussian mixture model (GMM) with $M=20$ bivariate normal distributions \cite{ha2017neural}  given by:
\vspace{-0.25cm}
\begin{equation}
p(\Delta x, \Delta y) = \sum_{j=1}^{M}\Pi_j \mathcal{N}(\Delta x, \Delta y\mid  \lambda_{j});  \; \sum_{j=1}^{M}\Pi_j= 1.
\end{equation}
\vspace{-0.4cm}

\noindent Each M bivariate normal distribution has \emph{five} parameters $\lambda = \{ \mu_{x}, \mu_{y}, \sigma_{x}, \sigma_{x}, \rho_{xy}\}$ with mean ($\mu_x, \mu_y$), standard deviation ($\sigma_x, \sigma_y$) and correlation ($\rho_{xy}$). The mixture weights of the GMM is modelled by a categorical distribution of size $\mathbb{R}^M$. Thus every time step's output $y_t$ modelled is of size  $\mathbb{R}^{5M + M + 3}$, which includes 3 logits for pen-state.  At time step $t$, a recurrent decoder network updates its state $s_t = (h_{t}, c_{t})$ as follows: $\mathrm{ s_t = RNN(s_{t-1} ; [g_{t}, P_{t-1}] )}$ where $g_{t}$ is the glimpse vector  encoding the information from specific relevant parts of the feature map $\mathcal{B}$ to predict $y_{t}$; $P_{t-1}$ is the last predicted point (start-token $P_0 =\{0, 0, 1, 0, 0\}$),  $[\cdot]$ signifies a concatenation operation. The glimpse/context vector is obtained by 2D attention as follows:

\vspace{-0.5cm}
\begin{align}
    \begin{cases}
      J =  \tanh(W_{\mathcal{B}} \circledast  \mathcal{B} + W_{S}h_{t-1});  \\ 
     \alpha_{i,j}  =  \mathrm{softmax}(W_{a}^T  J_{i,j})   \\
      g_{t} =  \sum_{i,j} \alpha_{i,j} \cdot \mathcal{B}_{i,j};  \; i=[1,..h'], \; j=[1,..w']
    \end{cases}
\end{align}
\vspace{-0.3cm}


\noindent where $W_{B}$, $W_{S}$, $W_{a}$ are the learnable weights. Calculating the attention weight $\alpha_{i,j}$ at every spatial position $(i,j)$, we employ a convolution operation ``$\circledast $" with $3\times3$ kernel $W_{\mathcal{B}}$ to consider the neighbourhood information in the 2D attention module, and $g_t$ is obtained by weighted summation operation at the end. A fully-connected layer over every hidden state outputs $y_t = W_yh_t + b_y$ , where $y_t \in \mathbb{R}^{6M+3}$. We refer to \cite{ha2017neural} for more details. Like the standard VAE, our generator $\mathcal{G}$ is trained from the weighted summation of a reconstruction loss ($\mathrm{L^R_{\mathcal{G}}}$) and a KL-divergence loss ($\mathrm{L^{kl}_{\mathcal{G}}}$)  with unit normal distribution as follows: 

\vspace{-0.5cm}
\begin{equation}\label{vae_loss}
\mathrm{L_{\mathcal{G}}^{vae} = L^{R}_{\mathcal{G}} + \omega_{kl} L^{kl}_{\mathcal{G}}},    
\end{equation}
\vspace{-0.5cm}


\noindent where  $L^{R}_{\mathcal{G}}$ is composed of the negative log-likelihood loss of the offsets $\Delta z = (\Delta x, \Delta y)$ and the pen states $(p_1, p_2, p_3)$: $\mathrm{L^{R}_{\mathcal{G}}} = - \frac{1}{T} \Big[ \sum_{i=1}^{T}\log p(\Delta z_i\mid \lambda_i) + \hat{p}_i\log(p_i) \Big]$.

\subsection{Baseline FG-SBIR Model} \label{sec:basemodel}
 
For the discriminative retrieval module $\mathcal{F(\cdot)}$, we use the state-of-the-art Siamese network \cite{song2017deep, dey2019doodle, bhunia2020sketch} (multi-branch with weight-sharing)  with soft spatial attention \cite{xu2015show} to focus on salient parts of the feature map. 
Concretely, given a photo or rasterized sketch image $I$, we use a pre-trained InceptionV3 model \cite{szegedy2016rethinking} to extract feature map $F' = f_{\theta}(I)$. This is followed by a residual connection between backbone feature and attention normalised feature to give $F = F' + F'\cdot f_{attn}(F')$, upon which global average pooling is performed to obtain final feature representation of size $\mathbb{R}^d$; and $f_{attn}$ is  modelled using 1x1 convolution with softmax across the spatial dimensions. For training, the distance to a sketch anchor (a) from a negative photo (n), denoted as $\beta^{-} = \left \| \mathcal{F}(a) - \mathcal{F}(n) \right \|_{2}$ should increase while that from the positive photo (p), $\beta^{+} = \left \| \mathcal{F}(a) - \mathcal{F}(p) \right \|_{2}$ should decrease. This is brought about by the triplet loss with a margin $\mu > 0$ as a hyperparameter:


\vspace{-0.3cm}
\begin{equation}\label{triplet}
\mathrm{L_{\mathcal{F}}^{trip} = max\{0, \mu + \beta^{+} - \beta^{-}\}}.
\vspace{-0.1cm}
\end{equation}

\subsection{Semi-Supervised Framework for FG-SBIR}
\vspace{-0.2cm}
\keypoint{Overview:} Firstly, we train the  photo-to-sketch generation model and discriminative fine-grained FG-SBIR model independently using the labelled training set $\mathrm{\mathcal{D}_{L}}$. Thereafter, through our semi-supervised learning framework,  $\mathcal{F}(\cdot)$ starts exploiting the unlabelled photos to improve its retrieval performance, while enhancing sketch generation quality of $\mathcal{G}(\cdot)$ by using $\mathcal{F}(\cdot)$ as a critic to provide training signal to the sketch-generation model $\mathcal{G}(\cdot)$ using both unlabelled and labelled photos simultaneously. Hence, both $\mathcal{G}(\cdot)$  and  $\mathcal{F}(\cdot)$ can now improve itself with the help of each other and by exploiting unlabelled photos (see Figure \ref{fig:Fig2}).  

\noindent \textbf{Certainty Score for Synthetic Photo-Sketch Pair:}  The generated photo-sketch pairs $\mathcal{D}_{U}'$ of unlabelled photos are sometimes noisy compared to real labelled photo-sketch pairs $\mathcal{D}_{L}$. This is mainly due to large possible output space \cite{song2018learning} of sketch drawing even with respect to a particular photo, as well as  difficulties in predicting the sketch ending token \cite{ha2017neural} in the sequential decoding process. Every synthetic photo-sketch instance pair needs to be handled individually based on their quality, thus requiring a specific \emph{certainty score} -- signifying the reliability of synthetic photo-sketch pair to train the retrieval model. Existing semi-supervised classification approach usually considers the probability distribution over classes to filter out noisy samples based on a predefined threshold \cite{yalniz2019billion}, top-K selection \cite{sohn2020fixmatch}, or uses entropy-based instance-wise weighting \cite{iscen2019label} to deal with noisy synthetic labels. 
A new solution is thus needed to not just measure the  quality of generated sketch itself, but to quantify how the generated sketch matches with the particular photo input, in order to help training the retrieval model. 

Inspired by the generative adversarial network \cite{chongxuan2017triple} where the sigmoid normalised output of discriminator shows the probability of being a real vs fake input sample, we use the \emph{discriminator's confidence} to quantify the quality of synthetic photo-sketch pairs. Specifically, the discriminator $\mathrm{D_{C}}$ learns to classify between real photo sketch pairs and generated pseudo photo-sketch pairs  (concatenated across channels). Thus, the learning objective for $\mathrm{D_{C}}$ is 

\vspace{-0.7 cm}
\begin{multline}\label{discri}
\mathrm{L_{D_C} = -\mathbb{E}_{({p_{L}; \,{s_{p,L}}}) \sim  \mathcal{D}_{L}} \big[\log \mathrm{D_{C}}\big(p_{L}, s_{p, L}\big)\big]}\\[-2pt]
\mathrm{-\mathbb{E}_{({p_{U};\, {s_{p,U}}}) \sim  \mathcal{D}_{U}'}\big[\log\big(1- \mathrm{D_{C}}\big(p_{U}, s_{p, U}\big)\big)\big]}.
\end{multline}
\vspace{-0.6cm}

\noindent This objective is computed via a binary cross-entropy loss using label $1$ for real pairs, and $0$ for synthetic ones. Thus, the discriminator's output  $\mathrm{\mathrm{D_{C}}(p_{U}, s_{p, U}) \in [0,1]}$ signifies the extent to which the synthetic photo-sketch pairs match with the distribution of real labelled photo-sketch pairs. Therefore, values closer to $1$ indicate better quality synthetic photo-sketch pairs. 

\keypoint{Tolerance against Noisy Pseudo-Labelled Data:}\label{NT}To further avoid over-fitting to noisy synthetic photo-sketch pairs, we introduce a consistency loss with respect to a pre-trained (on labelled dataset) retrieval model  as \emph{weak teacher} \cite{furlanello2018born}. More specifically, once the baseline FG-SBIR model is trained from labelled data, we keep a copy as $\mathcal{F}^T$ with weights frozen. As $\mathcal{F}^T$ has been trained from real clean photo-sketch pairs only, we expect that the feature embedding vector obtained from it would act as an additional supervision via distillation \cite{hinton2015kd} to regularise the main FG-SBIR model ($\mathcal{F}$) which is to be trained from both labelled  and synthetic photo-sketch pairs in a semi-supervised manner. This distillation process is expected to improve the tolerance against noisy information of synthetic data. Compared to cross-entropy loss \cite{hinton2015kd} used in distillation of classification network, a naive choice to design distillation for feature embedding network is to  minimise the distance between learnable student's embedding and teacher's embedding of a particular photo or sketch image individually. We term it \emph{absolute teacher}. However, instead of considering the actual embedding, we hypothesise that the \emph{relative distance} between paired photo and sketch, minimising which is the major purpose of embedding network, could be a better knowledge to be distilled. We term this as \emph{relative teacher}. Thus, given a photo-sketch image pair ($p, s_p$) and  $d(\cdot, \cdot)$ being a ${l_2}$ distance function,  the consistency loss for learnable student $\mathcal{F}$ with respect to pre-trained teacher $\mathcal{F}^T$ becomes as follows:  
\vspace{-0.25 cm}
\begin{equation}\label{eqn_3}
\mathrm{L}_{\mathcal{F}}^{KD}  =   \big\| d\big(\mathcal{F}^T(p),\mathcal{F}^T(s_p)\big) - d\big(\mathcal{F}(p),\mathcal{F}(s_p)\big) \big\|_{2}.
\end{equation}
\vspace{-0.75 cm}

\subsection{Joint Training} \label{jointtraining}
\vspace{-0.1 cm}
\keypoint{Optimising FG-SBIR Model:} We train the fine-grained SBIR model $\mathcal{F}$ using triplet loss over labelled photo-sketch pairs $\mathcal{D}_L$, \emph{instance weighted triplet loss} over generated pseudo photo-sketch pairs $\mathcal{D}_U$, and pre-trained teacher based consistency loss over both  $\mathcal{D}_L$ and $\mathcal{D}_U$. Given sampled data  $\mathcal{D}_L^i=\{p_L^i, s_{p,L}^i\} \sim \mathcal{D}_L$ and 
 $\mathcal{D'}_U^j=\{p_U^j, s_{p,U}^j\} \sim \mathcal{D}_U'$ (on \emph{same} ratio), the instance-wise weight is calculated as $\mathrm{\omega_{j} = D_C(\mathcal{D'}_U^j)}$. The overall semi-supervised loss to train the retrieval model becomes: 
 
\vspace{-0.6 cm}
\begin{equation}\label{train_retrieval}
\resizebox{0.90\hsize}{!}{%
$\mathrm{L}_{\mathcal{F}}^{all} =  L_{\mathcal{F}}^{trip}(\mathcal{D}_L^i) + \omega_j \cdot L_{\mathcal{F}}^{trip}(\mathcal{D'}_U^j)  + \lambda_{kd}  \cdot L_{\mathcal{F}}^{KD}(\mathcal{D}_L^i, \mathcal{D'}_U^j)$
}
\end{equation} 
\vspace{-0.6 cm}


\keypoint{Optimising Photo-to-Sketch Model:} Besides the fully-supervised VAE loss $L_{\mathcal{G}}^R$, during joint training, the photo-to-sketch generation model is also learned considering $\mathcal{F}$ and $\mathrm{D_C}$ as critics. In particular, if the generated sketch from $\mathcal{G}$ correctly depicts the corresponding input photo, the triplet loss for that generated photo-sketch pair from retrieval model would be low, signifying a better photo-sketch matching and generated sketch quality. Similarly, the higher the discriminator's output, the better the quality of generated photo-sketch pairs. However, these training signals from $\mathcal{F}$ and $\mathrm{D_C}$ cannot  be directly back-propagated to $\mathcal{G}$, as there exists a non-differentiable \emph{rasterization} operation $s_p$ before feeding the sketch-image to both retrieval model and discriminator.  Hence, we employ reinforcement learning  based on  policy-gradient  \cite{sutton2000policy}  with REINFORCE \cite{williams1992simple} deployed to estimate gradients with respect to parameters $\theta_{\mathcal{G}}$ of  $\mathcal{G}$ given some \emph{reward}. As $\mathcal{G}$ aims to lower this triplet loss value $L_{\mathcal{F}}^{trip}$ (Eqn.~\ref{triplet}), the reward should be negative of  $L_{\mathcal{F}}^{trip}$ that needs to be maximised. Similarly, the discriminator's output quantifying the goodness of photo-sketch pairs needs to be maximised. Thus the weighted joint reward is:
 
\vspace{-0.3cm}
\begin{equation}\label{reward}
\mathrm{\mathrm{R}_{\mathcal{G}} = -\lambda_{r1} \cdot L_{\mathcal{F}}^{trip}(\mathcal{D}^i) + \lambda_{r2} \cdot D_C(\mathcal{D}^i)}
\vspace{-0.1cm}
\end{equation}

\noindent where  $\mathcal{D}^i \sim \mathcal{D}_L \cup \mathcal{D}_U$. This reward could be computed for both labelled and unlabelled data as it does not need any ground-truth sketch-coordinates unlike the $L_{\mathcal{G}}^{vae}$ loss (Eqn.~\ref{vae_loss}). Thus two types of gradients are computed to update the parameter $\theta_{\mathcal{G}}$, one using policy gradient  \cite{sutton2000policy}  based on joint-reward guided by the retrieval model and the discriminator, and the other using back-propagation over only the labelled photos: 


\vspace{-0.5cm}

\begin{equation}\label{g_train}
\resizebox{1\hsize}{!}{
\begin{math}{
\begin{split}
 &\nabla_{\theta_{\mathcal{G}}}L(\theta_{\mathcal{G}}) = \underbrace{\nabla_{\theta_{\mathcal{G}}}L_{\mathcal{G}}^{vae}(\theta_{\mathcal{G}})}_\text{over only labelled data} \;\;\;\;\;\;\;\;\;\;\;\;\;\;\;\;\;\;\;\;\;\;\;\;\;\;\;\;\;\;\;\;\;\;\;\;\;\;\;\;\;\;\;\;\;\;\;\;\;\; \mathrm{(9)} \\[-13pt]
& \mathrm{-}\lambda_\mathcal{G}\sum_{i=1}^{T}\underbrace{\mathbb{E}_{\substack{p_i\sim p(q_i) \\ \Delta z_i\sim p(\Delta z_i \mid \lambda_i)}}\nabla_{\theta_{\mathcal{G}}} \Big( \log p(\Delta z_i \mid \lambda_i) + \log p(p_i) \Big) \cdot R_{\mathcal{G}}}_\text{over both labelled and unlabelled data (via policy gradient)} 
\nonumber
\end{split}}
\end{math}}
\end{equation}

\vspace{-0.4 cm}

\noindent In our experiments, we only update the final, fully-connected layer of sketch-decoder (with weights $W_y, b_y$ predicting 6M+3 outputs at every time step), at times using policy gradient, keeping rest of the parameters of $\mathcal{G}$ fixed. {We use a single global reward for the whole sketch-coordinate sequence, instead of local reward at every time step, that would otherwise need costly Monte Carlo roll-outs \cite{yu2017seqgan}}. Note that in our design, $\mathcal{G}$  and $\mathrm{D_C}$ are connected in a GAN-like fashion \cite{goodfellow2014generative} having adversarial objective. Moreover, the retrieval and generative models are trained alternatively improving each other over time (Algorithm \ref{algo}).  

\setlength{\textfloatsep}{0pt}
\begin{algorithm}[!hbt]
\sloppy
	\caption{Training of Semi-Supervised FG-SBIR}
	\begin{algorithmic}[1] 
        \State \textbf{Input}: Labelled photo-sketch pairs $\mathcal{D}_L$ and Unlabelled photos $\mathcal{D}_U$.  
        \label{algo}
        \State \textbf{Initialise hyper params}: $k_r$, $k_g$.
        \State \textbf{Pre-training}: $\mathcal{G}$ and $\mathcal{F}$ from $\mathcal{D}_L$ (using $\mathrm{L_{\mathcal{G}}^{vae}}$ \& $\mathrm{L_{\mathcal{F}}^{trip}}$). \cut{using $\mathrm{L_{\mathcal{G}}^{vae}}$ and $\mathrm{L_{\mathcal{F}}^{trip}}$ respectively.} 
        \While {not done training}
            \For {$k_{r}$ steps}
                \State Sample data $\mathcal{D}_L^{i} \sim \mathcal{D}_L$ and $\mathcal{D}_U^j \sim \mathcal{D}_U$.
                \State Get synthetic paired images $\mathcal{D'}_U^j$ using $\mathcal{G(\cdot)}$.
                \State \textsc{Train} $\mathcal{F}$ using $\{\mathcal{D}_L^i, \mathcal{D'}_U^j\}$ \Comment{Eqn.~\ref{train_retrieval}}
                \State \textsc{Train} $D_C$ using $\{\mathcal{D}_L^i, \mathcal{D'}_U^j\}$ \Comment{Eqn.~\ref{discri}}
            \EndFor
            \For {$k_{g}$ steps}
                \State Sample data $\mathcal{D}_L^{i} \sim \mathcal{D}_L$ and $\mathcal{D}_U^j \sim \mathcal{D}_U$.
                \State Get reward $R_{\mathcal{G}}$ using $\mathcal{F}$ and $D_C$.
                \State \textsc{Train} $\mathcal{G}$ using $\{\mathcal{D}_L^i, \mathcal{D}_U^j\}$ \Comment{Eqn.~\ref{g_train}}
            \EndFor
        \EndWhile
     \State \textbf{Output}: Optimised models $\mathcal{F}$, $\mathcal{G}$ and $\mathrm{D_C}$.
	\end{algorithmic}
\end{algorithm}

\vspace{-0.15cm}
\section{Experiments}\label{sec:experiments}
\vspace{-0.15cm}
\noindent \textbf{Datasets:} Two publicly available datasets, QMUL-Shoe-V2 \cite{pang2019generalising, riaz2018learning, song2018learning, bhunia2020sketch} and QMUL-Chair-V2 \cite{bhunia2020sketch, song2018learning} are used, which contain stroke-level coordinate information of sketches in addition to instance-wise paired sketch-photo labels, thus enabling us to train both retrieval and sketch-generative models. Out of the 6,730 sketches and 2,000 photos in Shoe-V2,  6,051 and 1,800 for training respectively, and the rest are for testing \cite{bhunia2020sketch, song2018learning}. The splits \cite{bhunia2020sketch, song2018learning} for Chair-V2 dataset are 1,275/725 sketches and 300/100 photos for training/testing respectively. In addition to these  labelled training data, we further use all 50,025 UT-Zap50K images \cite{yu2014fine} as unlabelled photos for shoe retrieval, and 7,800 unlabelled chair photos \cite{pang2020solving} are collected from shopping websites, including IKEA, Amazon and Taobao. Data, code, and models will be released soon.
\vspace{0.1cm} \\
\noindent \textbf{Implementation Details:}  Firstly, for sketch-generation, we use ImageNet pre-trained VGG-16 as encoder, excluding any global average pooling operation. We keep the  dimension ($N_z$) of $z$ as 128, the hidden state of the decoder LSTM as 512, the embedding dimension of the 2D-attention module as 256 respectively. \cut{A similar pre-training strategy to \cite{song2018learning} is adopted using rasterized sketches of QuickDraw dataset \cite{ha2017neural}.} We set the max sequence length to $100$, and the generative model is trained with a batch size of 64 with $\omega_{kl} = 1$ using the pre-training strategy from \cite{song2018learning}. Secondly, the retrieval model (ImageNet pre-trained Inception-V3 \cite{szegedy2016rethinking} ) is trained with a batch-size of 16 with a margin value of 0.3. Finally, after completing individual training from labelled data, we start \emph{joint training} (Section \ref{jointtraining}) by additionally exploiting unlabelled data using $k_g$ and $k_r$ as 5. Architecture of  $\mathrm{D_C}$ is from  \cite{isola2017image}. 
\cut{The discriminator used in the joint-training process has a similar design to \cite{isola2017image}.} 
We set $\lambda_{kd}=0.1$, $\lambda_{r1}=1$, $\lambda_{r2}=1$, and $\lambda_{\mathcal{G}}=10$ respectively. All images are resized to $256\times256$, with rasterization from sketch-coordinate involving a window of same size having centre scaling as well. We use Adam optimiser for both the generation and retrieval models with a learning rate of $0.0001$.

\vspace{0.1cm} 
\noindent \textbf{Evaluation Metric:} \textbf{(a) FG-SBIR:} Following existing FG-SBIR works \cite{yu2016sketch, pang2020solving}, we use $\mathrm{Acc@q}$, i.e. percentage of sketches having true-paired photo appearing in the top-q list. \textbf{(b) Sketch Generation:} Following \cite{song2018learning, sketchxpixelor}, sketch-generation is quantified from three perspectives (i) \emph{Recognition:} Using a ResNet-50 classifier trained on 250-classes from TU-Berlin sketch dataset, a generated sketch getting recognised as the same class as that of corresponding photo signifies category-level realism. (ii) \emph{Retrieval:}\footnote{Note: Retrieval accuracy is used to quantify both FG-SBIR and sketch generation performance. Please refer to \cite{song2018learning} for more details.} To judge whether the generated sketch has object-instance specific agreement, we check the retrieval accuracy $\mathrm{Acc@q}$ via a pre-trained FG-SBIR model using the generated sketches to retrieve corresponding photos of the testing set. (iii) \emph{Generation:} Following a recent sketch generation work \cite{sketchxpixelor}, we further calculate FID-score \cite{heusel2017gans} using a pre-trained sketch-classifier that captures both the quality and diversity of generated data compared to real human sketches. 
 
\vspace{-0.11cm} 
\subsection{Competitors}
\vspace{-0.20cm}
\keypoint{Sketch Generation:}  Sketch Generation could  be approached
in two following ways:  (a) \textit{Image-to-image translation} pipeline: \textbf{Pix2Pix} \cite{isola2017image} could be adapted to perform cross-modal translation in the image space. \textbf{PhotoSketch} \cite{li2019photo} extends further to handle the one-to-many possible nature of photo-conditioned sketch image generation problem, by calculating a mean loss over multiple sketches corresponding to a particular photo. (b) \textit{Image-to-sequence generation} pipeline:  \textbf{Pix2Seq} \cite{chen2017sketch} is the ablated version of our model having a convolutional encoder and LSTM decoder, without involving 2D-attention. \textbf{L2S} \cite{song2018learning} is an extension over \cite{chen2017sketch} that uses two-way cross domain translation with self-domain reconstruction for better regularisation.  \textbf{Ours-G} is a \emph{supervised} model with 2D-attention, trained independently from labelled data only. \textbf{Ours-G-full} is our final sketch-generative model involving joint-training to learn from both labelled and unlabelled  data. 

\keypoint{Fine-Grained SBIR:} We  compare with three groups of competitors. (a) \emph{state-of-the-art:} \textbf{SN-Triplet} \cite{yu2016sketch} employs triplet ranking loss with Sketch-a-Net as its baseline feature extractor. \textbf{SN-HOLEF} \cite{song2017fine}  is an extension over \cite{yu2016sketch}  employing spatial attention along with higher order ranking loss. \textbf{SN-RL} \cite{bhunia2020sketch} is a very recent work employing reinforcement learning based fine-tuning for on-the-fly retrieval. As early retrieval is not our objective, we cite result at sketch-completion point. (b) \emph{Exploiting unlabelled photos:}  There has been no prior work addressing semi-supervised learning for  FG-SBIR, and model designed for category-level retrieval \cite{jang2020generalized} does not fit here. We thus adopt a few works that could be used to leverage unlabelled photos. \textbf{Edgemap-Pretrain} \cite{radenovic2018deep} is a naive-approach to use edge-maps of unlabelled photos to pre-train the retrieval model.  While edge-maps hardly have any similarity to real free-hand sketches, they could be converted to better pseudo-sketches using the work \cite{riaz2018learning} that learns how to abstract sketches based on subset-stroke selection. We term it as \textbf{Edge2Sketch} \cite{riaz2018learning}. Recently, \textbf{Jigsaw-Pretrain}  \cite{pang2020solving} used jigsaw solving over the mixed patches between a particular photo (unlabelled) and its edge-map, as a pre-text task for self-supervised learning (SSL) to improve FG-SBIR performance. Furthermore, we term our self-implemented \emph{supervised} FG-SBIR model trained only on labelled data as  \textbf{Ours-F}. \textbf{Ours-F-Full} is our final retrieval model employing joint training over both labelled and unlabelled photos. We also replace our 2D-attention based sketch-generation process by baseline sketch-generative model Pix2Pix \cite{isola2017image} and L2S \cite{song2018learning}, and term them as \textbf{Ours-F-Pix2Pix} and \textbf{Ours-F-L2S}  respectively. Finally, we design a naive semi-supervised FG-SBIR baseline (\textbf{Vanilla-SSL-F}), where we \emph{blindly} (without instance-weighting and distillation) use the generated sketch to additionally train the retrieval model.

 

\setlength{\tabcolsep}{1.5pt}
\begin{table}[]
    \caption{Quantitative results of photo-to-sketch generation}
      \centering
    \scriptsize
    \begin{tabular}{cccccc}
        \midrule
        \multirow{2}{*}{Chair-V2} & \multicolumn{2}{c}{Recognition ($\uparrow$)} & \multicolumn{2}{c}{Retrieval($\uparrow$)} & \multirow{2}{*}{FID Score($\downarrow$)} \\
        \cmidrule{2-5}
         & $\mathrm{Acc.@1}$ & Acc.@10 & Acc.@1 & Acc.@10 & \\
        \midrule
        Pix2Pix \cite{isola2017image} & 4.5\% & 12.1\% & 2.4\% & 16.2\% & 33.4 \\
        PhotoSketch \cite{li2019photo} & 7.1\% & 14.3\% & 4.2\% & 17.9\% & 25.7 \\
        Pix2Seq \cite{chen2017sketch} & 5.4\% & 52.1\% & 4.0\% & 31.8\% & 14.5 \\
        L2S \cite{song2018learning} & 12.3\% & 53.8\% & 8.3\% & 36.7\% & 12.7 \\\hdashline
        Ours-G  (only labelled data)  & \blue{15.2\%} & \blue{56.9\%} & \blue{13.4\%} & \blue{40.7\%} & \blue{10.1} \\
        Ours-G-Full & \red{16.4\%} & \red{58.6\%} & \red{14.9\%} & \red{42.6\%} & \red{8.9} \\
        \toprule
        \multirow{2}{*}{Shoe-V2} & \multicolumn{2}{c}{Recognition($\uparrow$)} & \multicolumn{2}{c}{Retrieval($\uparrow$)} & \multirow{2}{*}{FID Score ($\downarrow$)} \\
        \cmidrule{2-5}
         & Acc.@1 & Acc.@10 & Acc.@1 & Acc.@10 & \\
        \midrule
        Pix2Pix \cite{isola2017image} & 6.2\% & 14.5\% & 1.8\% & 8.4\% & 31.7\% \\
        PhotoSketch \cite{li2019photo} & 8.9\% & 17.3\% & 3.4\% & 10.2\% & 24.3\% \\
        Pix2Seq \cite{chen2017sketch} & 51.3\% & 86.6\% & 5.1\% & 25.8\% & 11.3 \\
        L2S \cite{song2018learning} & 53.7\% & 89.7\% & 6.2\% & 28.6\% & 10.7 \\\hdashline
        Ours-G (only labelled data) & \blue{56.3\%} & \blue{91.9\%} & \blue{9.7\%} & \blue{33.6\%} & \blue{9.5} \\
        Ours-G-Full & \red{58.1\%} & \red{93.4\%} & \red{12.3\%} & \red{35.4\%} & \red{8.3} \\
        \midrule
    \end{tabular}
    \label{tab:generation}
     \vspace{0.1cm}
\end{table}

\vspace{-0.1cm}
\subsection{Performance Analysis }\label{perm}
 \vspace{-0.1cm}

\begin{figure*}[!h]
\begin{center}
  \includegraphics[width=\linewidth]{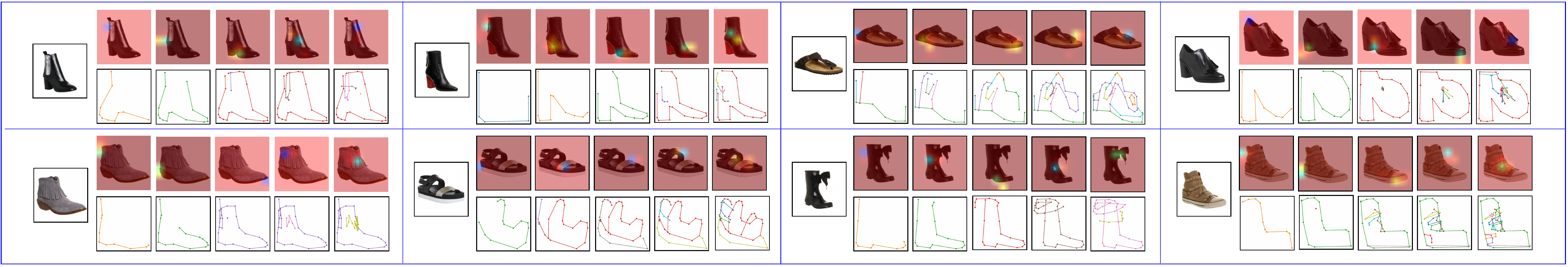}
\end{center}
\vspace{-.20in}
\caption{Qualitative results on our photo-to-sketch generation process, where sketch is shown with attention-map at progressive instances.}
\vspace{-.18in}
\label{fig:ablation2}
\end{figure*}

\noindent \textbf{Photo-to-Sketch Generation:} From Table \ref{tab:generation}, we observe: \textbf{(i)} \emph{Pix2Pix} and \emph{PhotoSketch}  based on cross-modal translation in pixel space perform poorly. They fail to capture the abstraction in human sketching style, where distribution gap with real sketches is reflected in their significantly poor FID scores. \textbf{(ii)} \emph{Pix2Seq} and \emph{L2S} outputs vector sketches by sequentially predicting sketch coordinates, thus possessing higher similarity towards human sketches. 
They however, still lag behind our ablated version \emph{Ours-G} in scores.
As both of them reduce the spatial dimension of the convolutional feature-map to a global context vector, spatial information is significantly compromised, with the decoder receiving little guidance from the vector on exact drawing content. In contrast, we retain the spatial dimension of feature-map, and employ 2D-attention to focus on that specific part of the photo it draws at any time step.  \textbf{(iii)} \emph{L2S} is a notably close competitor to ours in terms of recognition accuracy, but better information passage between every time step of decoder and convolutional encoder delivers much better sketches with fine-grained details (reflected by retrieval accuracy).  Furthermore, our final model \textit{Ours-G-Full} employs joint training with a retrieval model to additionally exploit the unlabelled photos, improving sketch generation performance  (retrieval Acc@1) from $9.7\%$ to $12.3\%$ by $2.6\%$ over our baseline \emph{Ours-G} on Shoe-V2, thus justifying the benefits of our semi-supervised learning. Some qualitative results are shown in Figure \ref{fig:ablation2}. Blue denotes a supervised baseline, while red is \textit{Ours-G(F)-Full}.
\vspace{0.1cm}\\
\setlength{\tabcolsep}{6pt}
\begin{table}[]
    \centering
    \caption{Quantitative results of fine-grained SBIR}
    \scriptsize
    \begin{tabular}{ccccc}
        \hline
        \multirow{3}{*}{Methods} & \multicolumn{2}{c}{Chair-V2} & \multicolumn{2}{c}{Shoe-V2} \\
         \cmidrule{2-5}
         & Acc.@1 & Acc.@10 & Acc.@1  & Acc.@10 \\
        \midrule 
        SN-Triplet \cite{yu2016sketch} & 47.4\% & 84.3\% & 28.7\% & 71.6\%  \\
        SN-HOLEF \cite{song2017fine} & 50.7\% & 86.3\% & 31.2\% & 74.6\%  \\
        SN-RL \cite{bhunia2020sketch} & 51.2\% & 86.9\% & 30.8\% & 74.2\%\\
        \hdashline
        Edgemap-Pretrain \cite{radenovic2018deep} & 53.9\% & 87.7\% & 33.8\% & 80.9\% \\
        Edge2Sketch-Pretrain \cite{riaz2018learning} & 54.3\% & 88.2\% & 34.2\% & 81.2\% \\
        Jigsaw-Pretrain \cite{pang2020solving} & 56.1\% & 88.7\% & 36.5\% & 85.9\% \\
        \hdashline
        Ours-F  (only labelled data)  & \blue{53.3\%} & \blue{87.5\%} & \blue{33.4\%} & \blue{80.7\%} \\
        Vanilla-SSL-F  & 49.6\% & 85.6\% & 30.6\% & 74.3\% \\
        Ours-F-Pix2Pix  & 53.2\% & 87.5\% & 33.2\% & 80.1\%  \\
        Ours-F-L2S & 57.6\% & 89.4\% & 36.6\% & 84.7\%  \\
        Ours-F-Full & \red{60.2\%} & \red{90.8\%} & \red{39.1\%} & \red{87.5\%} \\
        \hline
    \end{tabular}
    \label{tab:retrieval}
    \vspace{0.4cm}
\end{table}
\noindent \textbf{Fine-grained SBIR:} From Table \ref{tab:retrieval}, we observe: \textbf{(i)} Our baseline retrieval model is noticeably better than \emph{SN-Triplet}, and lies at par with recent state-of-the-art FG-SBIR baselines like \emph{SN-HOLEF} and \emph{SN-RL}. \textbf{(ii)} With regards to exploiting unlabelled photos, \emph{Edgemap-Pretrain} offers marginal improvement while using it on top of our baseline with ImageNet pretrained weights. Aligning with the intuition, while edge-maps are further augmented with \emph{Edge2Sketch} by a subset of stroke selection to model the abstracted nature of sketch over edge-maps, it increases retrieval performance by a reasonable margin. In context of using edge-maps for pre-training, \emph{Jigsaw-Pretrain} provides maximum benefits, but still lags behind our final model \emph{Ours-F-Full}. \textbf{(iii)} While edge-map does not posses sketch abstraction knowledge of human sketching style, our approach of using a sequential photo-to-sketch generation model to generate synthetic photo-sketch pairs for unlabelled photo encodes better knowledge to enhance generalisation.  However, it is noteworthy that  \emph{Vanilla-SSL-F} blindly using synthetic sketch-photo pairs yields performance lower than the supervised one due to overfitting on noisy information. Overall, for fine-grained SBIR, due to our proposed semi-supervised learning, the retrieval accuracy $\mathrm{Acc@1}$ of \emph{Ours-F-Full } increases from   $33.4\%$ to $39.1\%$ by a margin of $5.7\%$ over our baseline \emph{Ours-F} on Shoe-V2. Moreover, replacing our photo-to-sketch generation model by \emph{L2S} and \emph{Pix2Pix} reduces  the same by $2.5\%$  and $5.9\%$ respectively, thus justifying the importance of our sketch generative model with 2D attention.  \textbf{(iv)} Note that policy-gradient based RL scheme could be avoided by using Pix2Pix for sketch generation, and gradient can directly be back-propagated from retrieval to generative model. However, that is still found to be inferior to ours.

\vspace{-0.1cm} 
\subsection{Ablation Study}
\begin{figure} 
    \centering
    \includegraphics[width=0.49\linewidth, height=3.4cm]{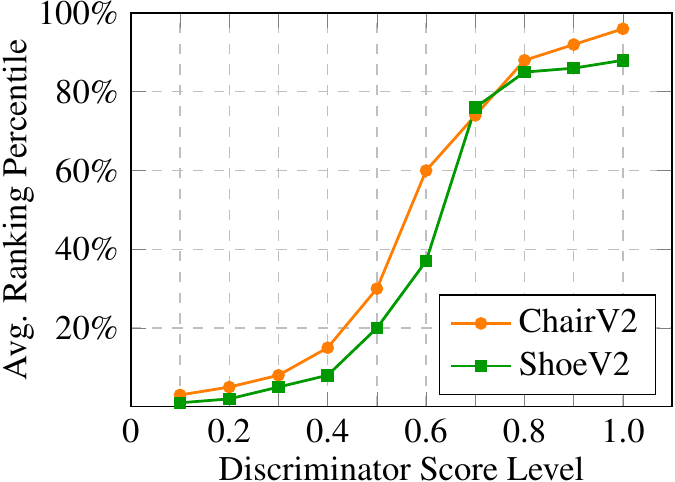}
    \includegraphics[width=0.49\linewidth, height=3.4cm]{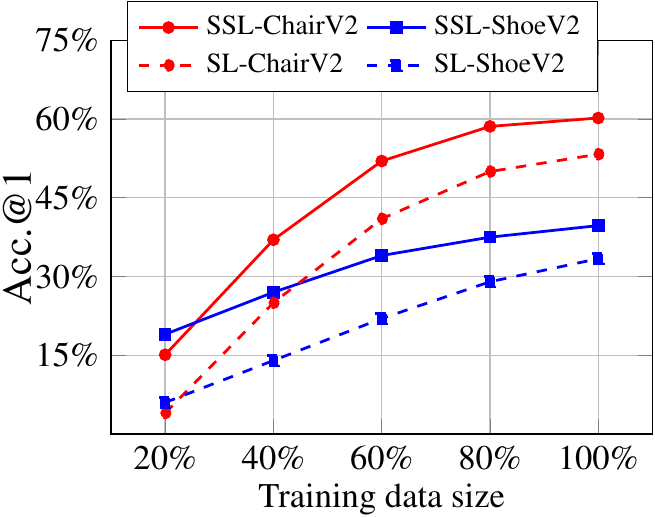}
    \caption{(a): Consistency of discriminator's certainty score. (b): Varying training data size for FG-SBIR - Semi-Supervised Learning (SSL) vs Supervised-Learning (SL).}
    \label{fig:ablation3}
    \vspace{0.2cm}
\end{figure}

 \vspace{-0.09cm}
\label{abla}
A thorough ablative study on Shoe-V2 dataset verifies contributions of individual design components in Table \ref{tab:ablation}. \textbf{{\emph{[i]} Instance weighting for retrieval}:} To simply judge the contribution of discriminator ($\mathrm{D_C}$) guided instance weighting we remove it, and adapt the framework accordingly. Consequently $\mathrm{Acc@1}$ retrieval performance significantly drops to $36.8\%$ with a decrease of $2.3\%$ on Shoe-V2. Due to sigmoid normalisation \cite{chongxuan2017triple}, the output of $\mathrm{D_C}$ falls in [0,1]. We quantify it as 10 discrete levels with a step size $0.1$. We calculate the average ranking percentile (ARP) of synthetic sketch-photo pairs from testing set which fall under the same discrete level, and plot it against $10$ different levels. From Figure \ref{fig:ablation3} (a), it is evident that the synthetic sketch-photo pairs having higher discriminator score (towards $1$) tend to have much better ARP \cite{bhunia2020sketch} values (i.e better quality), while those with lesser ARP values are assigned with lesser (towards $0$) certainty score by the discriminator. This observation is consistent with our assumption that $\mathrm{D_C}$ should quantify quality of synthetic sketch-photo pairs for instance-wise weighting.  \textbf{{\emph{[ii]}  Distillation based noise tolerance for retrieval}:}  Removing knowledge distillation based regularisation, which additionally tries to provide tolerance against noisy synthetic sketch-photo pairs, $\mathrm{Acc@1}$ is decreased by $1.8\%$ to $37.3\%$ on Shoe-V2 dataset. Our  \emph{relative teacher} based distillation process (Section \ref{NT}) for retrieval network surpasses \emph{absolute teacher} alternative by a margin of $0.9\%$ ($\mathrm{Acc@1}$) on Shoe-V2, thus confirming its usefulness. \textbf{{\emph{[iii]} 2D-attention for sketch-generation}:} The use of 2D attention significantly improves the sketch generation performance, providing better fine-grained agreement with the input photo. While we employ a 3x3 convolutional kernel to aggregate neighbourhood information, using an 1D attention that treats feature maps as 1D sequence, the retrieval accuracy $\mathrm{Acc@1}$ of generated sketches on Shoe-V2 drops to $8.1\%$ by margin of $4.2\%$. We conjecture that 2D-spatial attention has higher efficiency in generating fine-grained sequential sketches from input photo than two-way translation based regularisation as done in \emph{L2S} \cite{song2018learning}.  
\textbf{{\emph{[iv]} Significance of joint-training}:} (a) A direct way of judging efficiency of joint training is employing separately trained photo-to-sketch generation model to augment synthetic sketch-photo pairs, and using them \emph{blindly} to train the retrieval model along with labelled data \emph{without} instance weighting or teacher-regularisation.  This however lags behind the baseline (supervised) fine-grained SBIR model by $2.8\%$ (i.e. $30.6\%$), as the model over-fits to the noisy information present in synthetic sketch-photo data. This confirms that naively using sketch generation does not help at all. (b) The retrieval accuracy ($\mathrm{Acc@1}$) from sketch generation performance improves by $1.4\%$ with an additional policy-gradient based training taking reward (Eqn.~\ref{reward}) from the retrieval model and the discriminator $\mathrm{D_C}$ as critic. Individually, they help to improve by $0.9\%$ and $0.8\%$, respectively under the same metric.  (c) Furthermore, we compute the performance of our semi-supervised framework at varying training data size for both processes in Figure \ref{fig:ablation3} (b). We notice a significant overhead compared to our supervised baseline model for each dataset individually.

\vspace{-0.2cm} 
\setlength{\tabcolsep}{4.5pt}
\begin{table}[!hbt]
    \centering
    \caption{Ablative study on Shoe-V2: Instance Weighting (IW), Teacher Regularisation (TR), Attention (AT), Joint-Training (JT).}
    \scriptsize
    \begin{tabular}{cccccccc}
        \hline
        \multirow{3}{*}{IW} & \multirow{3}{*}{TR} & \multirow{3}{*}{AT} & \multirow{3}{*}{JT} & \multicolumn{2}{c}{Fine-Grained SBIR} & \multicolumn{2}{c}{Sketch Generation} \\
        \cline{5-8}
         & & & & \multirow{2}{*}{Acc.@1} & \multirow{2}{*}{Acc.@10} & Recognition & Retrieval \\
         \cline{7-8}
         & & & & & & Acc.@1 & Acc.@1 \\
        \hline
        \cmark & \cmark & \cmark & \cmark & 39.1\% & 87.5\% & 58.1\% & 12.3\% \\
        \hline
        \xmark & \cmark & \cmark & \cmark & 36.8\% & 85.4\% & 57.3\% & 11.2\% \\
        \cmark & \xmark & \cmark & \cmark & 37.3\% & 86.1\% & 57.8\% & 12.1\% \\
        \cmark & \cmark & \xmark & \cmark & 37.6\% & 86.1\% & 51.3\% & 5.1\% \\
        \cmark & \cmark & \cmark & \xmark & 37.9\% & 86.6\% & 56.3\% & 9.7\% \\
        \hline
        \xmark & \xmark & \xmark & \xmark & 31.1\% & 75.4\% & 51.3\% & 5.1\% \\
        \hline
    \end{tabular}
    \label{tab:ablation}
    \vspace{-0.2cm}
\end{table}%

\vspace{-0.6cm}
\section{Conclusion}
\vspace{-0.1cm}
We have proposed a semi-supervised fine-grained sketch-based image retrieval framework to solve the data scarcity problem. To this end, we proposed to treat sequential photo-to-sketch generation and fine-grained sketch-based image retrieval as two conjugate problems along with various regularizers to address the intricate issues of reliability and tolerance to noisy synthetic sketch-photo pairs. This leads to substantial improvement on existing baselines in sparse data-scenarios for FG-SBIR.

{\small
\bibliographystyle{ieee_fullname}
\bibliography{Original_egbib}
}

\cleardoublepage 
 \appendix

\renewcommand{\thesubsection}{\Alph{subsection}}
\setcounter{figure}{0}
\setcounter{table}{0}

 \onecolumn{%
  \centering
 \Large \bf Supplementary material for \\ More Photos are All You Need: Semi-Supervised Learning for Fine-Grained Sketch Based Image Retrieval \par
 \vspace{0.5cm}
 }

\section{Necessity for rasterization}
This is common practice in the FG-SBIR literature. Rasterized sketch-images tend to have better spatial encoding than coordinate-based alternatives \cite{pang2019generalising, bhunia2020sketch}. This is also verified in our work -- removing rasterization and using sketch-coordinate retrieval reduces acc@1 to only 7.6\% on Shoe-V2. 

\section{Motivation behind using discriminator for certainty score:}

 We are mostly inspired by recent image generation works \cite{shama2019adversarial, huh2019feedback} that use the discriminator scores to iteratively improve generation quality. We also did an ablative study to investigate further (see L768-785 and Fig. 4(a)). We found that synthetic sketch-photo pairs having higher discriminator score, tend to have much better quality, and vice-versa. We will add some qualitative examples to further illustrate this correlation in supplementary materials. Defining a hard threshold (optimised) to eliminate bad generated sketches is an option -- new experiments show acc@1 lags by around 2\% compared to ours on Shoe-V2.

\section{More details on experimental setup and analysis:}

(i) Our self-designed baselines use the same backbone network, while joint-training is employed for Ours-F-Pix2Pix, Ours-F-L2S and Ours-F-Full. 

\vspace{0.1cm}
(ii) SOTA data-augmentation strategies are already adopted by existing FG-SBIR works \cite{pang2019generalising, BMVC_hierarchy}. However, they fail to capture the significant style variations that exist in real human sketches. In fact we already compare with \cite{yu2016sketch} which employed such sketch specific augmentation strategies, and it is found to be much inferior to our semi-supervised framework (see Table. 2). 

\vspace{0.1cm}
(iii)  Optimising the final layer  (using Eq. 9 in our case) is a very common practice during fine-tuning with RL, and is heavily adopted by the image-captioning literature \cite{gao2019self}, and very recently by on-the-fly FG-SBIR \cite{bhunia2020sketch}.

\vspace{0.1cm}
(iv) Edge-map hardly resembles the highly abstracted and subjective nature of amateur human sketches. For example, sketches do not follow the perfect edge boundary unlike edge-maps, thus model trained on pseudo-sketches via edge2sketch \cite{riaz2018learning} falls short to generalise to real human sketches. 

\vspace{0.1cm}
(v) Acc@1 without using RL scheme for Ours-F-Pix2Pix is $34.14\%$.

\vspace{0.1cm}
(vi) In future, our photo-to-sketch generation model could further be evaluated on Sketchy \cite{sangkloy2016sketchy}, however, it seems to be comparatively difficult than that of QMUL-ShoeV2 due to more noisy background.


\end{document}